\documentclass[letterpaper, 10 pt, conference]{ieeeconf}  
\IEEEoverridecommandlockouts                              
\usepackage{amsmath,amssymb,amsfonts}
\usepackage{graphicx}
\usepackage{textcomp}
\usepackage{xcolor}
\usepackage{subcaption}
\usepackage{booktabs}
\usepackage{tabularx}
\usepackage{array}
\usepackage{xurl}
\usepackage{tikz}
\usepackage{pgfplots}
\usepackage{fancyhdr}

\fancypagestyle{firstpage}{%
  \lhead{Accepted to American Control Conference (ACC) 2024}
  \rhead{}
}
\title{\LARGE \bf
A Comparison Between Lie Group- and Lie Algebra- Based Potential Functions for Geometric Impedance Control 
}

\author{Joohwan Seo$^{1}$, Nikhil Potu Surya Prakash$^{1}$, Jongeun Choi$^{1,2}$ and Roberto Horowitz$^{1}$
\thanks{This research is partially funded by (1) the Tsinghua-Berkeley Shenzhen Institute (TBSI) phase II and (2) the Hong Kong Center for Construction Robotics Limited (HKCRC).
Jongeun Choi was supported by the National Research Foundation of Korea (NRF) grants funded by the Korea Government (MSIT) (No.RS-2023-00221762)}
\thanks{$^{1}$Department of Mechanical Engineering, University of California, Berkeley
        {\tt\small \{joohwan\_seo, nikhilps, horowitz\}@berkeley.edu}}%
\thanks{$^{2}$School of Mechanical Engineering, Yonsei University
        {\tt\small jongeunchoi@yonsei.ac.kr}}%
}

\newcommand{\CLG}{\text{GIC-1}}
\newcommand{\CLA}{\text{GIC-2}}

\definecolor{color0}{rgb}{0.8235,0,0} 
\definecolor{color1}{rgb}{0.07843,0.549,0.07843} 
\definecolor{color2}{rgb}{0,0,1} 
\definecolor{color3}{rgb}{1,0.5137,0.4824} 
\definecolor{color4}{rgb}{0.5098,0.8588,0.1961} 
\definecolor{color5}{rgb}{0.4314,0.7451,0.9804} 
\definecolor{color6}{rgb}{0.3451,0.3451,0.3451} 
\definecolor{color7}{rgb}{0.6863,0.6863,0.6863} 
\definecolor{color8}{rgb}{0,0,0} 
\definecolor{color9}{rgb}{0.75,0.25,0} 
\definecolor{color10}{rgb}{0,0.8,0} 
\definecolor{color11}{rgb}{0.44, 0, 0.8 } 

\newcommand{\plotLine}{1.25pt}
\newcommand{\plotDotted}{1.5pt}

\newcommand{\tr}{\text{tr}}
\newcommand{\so}{\mathfrak{so}(3)}
\newcommand{\se}{\mathfrak{se}(3)}
\newcommand{\SO}{SO(3)}
\newcommand{\SE}{SE(3)}

\newcommand{\Ad}{\text{Ad}}
\newcommand{\ad}{\text{ad}}

\newcommand{\fg}{f_{_G}}
\newcommand{\fgone}{f_{_{G,1}}}
\newcommand{\fgtwo}{f_{_{G,2}}}

\newcommand{\psin}{\|\psi\|}

\newcommand{\gfrak}{\mathfrak{g}}

\newtheorem{theorem}{\textbf{Theorem}}
\newtheorem{remark}{\textbf{Remark}}

\newtheorem{assumption}{\textbf{Assumption}}

\begin{document}

\maketitle
\thispagestyle{empty}
\pagestyle{empty}

\begin{abstract}
In this paper, a comparison analysis between geometric impedance controls (GICs) derived from two different potential functions on $\SE$ for robotic manipulators is presented. The first potential function is defined on the Lie group, utilizing the Frobenius norm of the configuration error matrix. The second potential function is defined utilizing the Lie algebra, i.e., log-map of the configuration error. Using a differential geometric approach, the detailed derivation of the distance metric and potential function on $\SE$ is introduced. The GIC laws are respectively derived from the two potential functions, followed by extensive comparison analyses. In the qualitative analysis, the properties of the error function and control laws are analyzed, while the performances of the controllers are quantitatively compared using numerical simulation. 
\end{abstract}

\thispagestyle{firstpage}
\section{Introduction}\label{Sec:1}
Following its initial introduction in \cite{hogan1985impedance}, impedance control has emerged as the predominant method for controlling the position of an end-effector, especially in scenarios involving frequent interactions with the environment \cite{abu2020variable}. This control strategy enables robots to safely engage with their environments by attaining the desired impedance in the form of a mass-spring-damper system, without explicitly exerting force on the environment, as opposed to hybrid position-force control \cite{suomalainen2022survey}.

The configuration of a manipulator's end-effector is intricately defined by its position and orientation, effectively residing within the Special Euclidean group SE(3). The investigation of the geometric properties of manipulators dates back to the early stages of robotics \cite{park1995distance,park1995lie}, and has since become a standard approach to analyze manipulator kinematics using geometric principles \cite{lynch2017modern, murray1994mathematical}.
Nevertheless, to date, the majority of impedance control schemes have tended to address translation and orientation dynamics as distinct entities. This dichotomy can be observed not only in prior works like \cite{caccavale1999six} but also in more recent works such as \cite{ochoa2021impedance, shaw2022rmps, jeong2022memory, zhang2021learning}, 
implying a potential for further performance improvement by considering the inherent geometric structure of these systems.

The $\SO$ and $\SE$ have been major research interests within the domain of geometric control. This focus stems from the fact that classical mechanical systems are often regarded as rigid bodies, and $\SO$ and $\SE$ offer powerful mathematical tools for effectively representing such rigid-body motions. The PD control designs on these groups have been designed in \cite{bullo1995proportional,bullo1999tracking} employing the principles of differential geometry. Specifically, the geometric control design on $\SO$ was revisited in \cite{lee2010geometric} to the Unmanned Aerial Vehicle (UAV) application and widely utilized in the UAV field thereafter \cite{sreenath2013geometric, lee2013nonlinear}.


A well-known limitation of the control design in \cite{lee2010geometric} is its tendency for the error vector's magnitude to diminish, even when the configuration error remains large \cite{lee2012exponential}. This issue of slow convergence has been attributed to the ill-shaped Lie group-based potential function in \cite{teng2022lie}. 
To overcome such drawbacks, the Lie algebra-based potential function formulation was proposed as an alternative in \cite{teng2022lie}, where its fundamental concept of utilizing Lie algebra originates from \cite{bullo1995proportional, park1995distance}. 


In our previous work \cite{seo2023geometric}, the geometric impedance control (GIC) was proposed for manipulator control on $\SE$. In particular, the control design scheme of \cite{lee2010geometric} on $\SO$ was extended to $\SE$, and was applied to the manipulator system by combining with operational space formulation \cite{khatib1987unified}. Moreover, 
when GIC is combined with a learning variable impedance control framework, 
the learning transferability of manipulation tasks could be achieved by leveraging $\SE$ invariance and equivariance \cite{seo2023contact}. 

In this paper, we revisit geometric control on $\SE$, demonstrating its applicability to the robotic manipulator control design.
Specifically, we focus on providing a comparative exploration of two pivotal approaches for selecting potential functions and the resultant control laws: the Lie group-based approach and the Lie algebra-based approach. The primary contributions of this paper can be summarized as follows:
\begin{enumerate}
    \item A review of the distance metrics and potential functions on $\SE$ is provided.
    \item We provide the control laws and stability analysis derived from a choice of potential functions.
    \item We provide extensive comparison analyses between two control laws so that readers can be fully aware of the design choices. 
\end{enumerate}

\section{Backgrounds}\label{Sec:2}
In this section, we provide a brief description of the Lie group and Lie algebra that naturally arise in robotics applications. In particular, as the workspace of the manipulator lies on $\SE$, we focus on the formulation on $\SE$ and $\SO$.
\subsection{Lie group and Lie algebra of $\SO$ and $\SE$}
We first denote $G \subset \SE$ be a matrix Lie group and $\gfrak \subset \se$ as its Lie algebra. We note that we follow standard notations of hat map and vee map presented in \cite{seo2023geometric, seo2023contact} and refer to \cite{murray1994mathematical, bullo1995proportional} for the details of $\SE$ and $\se$ formulations. A dynamical system with a state given by $g \in G$ evolves as:
\begin{equation}
    \dot{g} = g \hat{V}^b, \quad \hat{V}^b \in \gfrak
\end{equation}
where ${V}^b \in \mathbb{R}^6$ denotes a velocity in body-frame. Note that we are only interested in body-frame velocity formulation since the distance metrics in Section.~\ref{Sec:3} are left-invariant; thus, it is natural to design in body-frame coordinate as proposed in \cite{seo2023geometric,park1995distance}. We also use the notation $V^b$ and $V$ interchangeably. 

Using homogeneous matrix representations, elements on $\SE$ and $\se$ can be denoted by
\begin{equation} \label{eq:SE(3)_and_se(3)}
    g = \begin{bmatrix}
        R & p \\ 
        0 & 1
    \end{bmatrix}, \quad 
    \hat{V} = \begin{bmatrix}
        \hat{\omega} & v \\
        0 & 1
    \end{bmatrix},
\end{equation}
where $g = (R,p)\in \SE$ and $\hat{V} = (\hat{\omega}, v) \in \se$ are also used for compact notation. For the column vector representation of $V$, we follow the convention of \cite{murray1994mathematical} and put translational components on the upper side and rotational parts on the lower side as $V = [v^T, \omega^T]^T$.

The Adjoint map $\Ad_g$ and adjoint map $\ad_{V}$ are originally defined on the $\se$ domain, i.e., $\Ad_g, \ad_{V}: \gfrak \to \gfrak$. However, it can be interpreted as linear mappings acting on the velocity $V \in \mathbb{R}^6$ in column vector representation. We are interested in such linear mapping representations, which are denoted as follows \cite{bullo1995proportional}.
\begin{equation} \label{eq:Ad_and_ad}
    \Ad_g = \begin{bmatrix}
        R & \hat{p} R \\
        0 & R
    \end{bmatrix}, \quad \ad_{V} = \begin{bmatrix}
        \hat{\omega} & \hat{v} \\
        0 & \hat{\omega}
    \end{bmatrix}
\end{equation}
The exponential map and logarithm map connect the Lie group and Lie algebra. In particular, for $\SE$ and $\se$, the exponential mapping $\exp: \gfrak \to G$ is defined as follows:
\begin{align} \label{eq:exp_mapping}
    \exp_{\SO}(\hat{\psi}) &= I + \sin{\psin}  \dfrac{\hat{\psi}}{\psin} + (1 - \cos{\psin}) \dfrac{\hat{\psi}^2}{\psin^2} \nonumber\\
    \exp_{\SE}(\hat{\xi}) &= \begin{bmatrix}
        \exp_{\SO}(\hat{\psi}) & A(\psi) b\\
        0 & 1
    \end{bmatrix},   
\end{align}
where $\hat{\psi} \in \so$, $\hat{\xi} = (\hat{\psi}, b) \in \se$, $\|\cdot\|$ is a standard Euclidean norm, together with
\begin{equation} \label{eq:A_mat}
    A(\psi) \!=\! I \!+\! \left(\dfrac{1 \!-\! \cos{\psin}}{\psin}\right)\!\dfrac{\hat{\psi}}{\psin} \!+\! \left(1 \!-\! \dfrac{\sin{\psin}}{\psin}\right)\!\dfrac{\hat{\psi}^2}{\psin^2}.
\end{equation}
Note that the first equation of \eqref{eq:exp_mapping} is known as Rodrigues' formula. $\hat{\xi} = \log{(g)} \in \gfrak$ is now can be defined as the exponential coordinates of the group element $g$, and the logarithmic map is regarded as a local chart of the manifold $G$ \cite{bullo1995proportional}. The logarithm map on $g = (R,p) \in \SE$ such that $\tr(R) \neq -1$ is defined in the following way \cite{bullo1995proportional}.
\begin{equation} \label{eq:log_mapping_SE}
    \begin{split}
        \log_{\SE}(g) &= \begin{bmatrix}
        \hat{\psi} & A^{-1}(\psi) p\\
        0 & 0
        \end{bmatrix} \in \se, \quad \text{with}\\
        A^{-1}(\psi) &= I - \dfrac{1}{2}\hat{\psi} + \left(1 - \alpha(\psin)\right) \dfrac{\hat{\psi}^2}{\psin^2},
    \end{split}
\end{equation}
where $\hat{\psi} = \log_{\SO}(R)$, $\alpha(y) = (y/2)\cot{(y/2)}$. In addition,
\begin{equation} \label{eq:log_mapping_SO}
    \log_{\SO}(R) = \dfrac{\phi}{2\sin{\phi}}(R - R^T) \in \so,
\end{equation}
with $\cos{\phi} = \tfrac{1}{2}(\tr(R) - 1)$ and $|\phi| < \pi$. 

\subsection{$\SE$ Operational Space Formulation}
We formulate the manipulator dynamics by considering $\SE$ as an operational space. Employing the operational space formula from \cite{khatib1987unified}, the manipulator dynamics using the body-frame velocity $V^b$ is given by:
\begin{align} \label{eq:robot_dynamics_eef}
    \hspace{-5pt}\tilde{M}(q)\dot{V}^b &+ \tilde{C}(q,\dot{q})V^b + \tilde{G}(q) = \tilde{T} + \tilde{T}_e, \text{ where}\\
    \tilde{M}(q)&= J_b(q)^{-T} M(q) J_b(q)^{-1}, \nonumber\\
    \tilde{C}(q,\dot{q})&=J_b(q)^{-T}(C(q,\dot{q}) \!-\! M(q) J_b(q)^{-1}\dot{J})J_b(q)^{-1}, \nonumber\\
    \tilde{G}(q)&= J_b(q)^{-T} G(q),\; \tilde{T}= J_b(q)^{-T} T,\; \tilde{T}_e= J_b(q)^{-T} T_e.\nonumber
\end{align}
where $J_b$ is a body-frame Jacobian matrix which relates body velocity and joint velocity by $V^b = J_b \dot{q}$.
The original manipulator dynamics derived from the Lagrangian in joint space is given as
\begin{equation}\begin{aligned} \label{eq:robot_dynamics}
    M(q)\ddot{q} + C(q,\dot{q})\dot{q} + G(q) = T + T_e,
\end{aligned}\end{equation}
where $q\in \mathbb{R}^n$ is joint coordinate, $M(q)\!\in\!\mathbb{R}^{n\times n}$ is the symmetric positive definite inertia matrix, $C(q,\dot{q})\!\in\!\mathbb{R}^{n \times n}$ is a Coriolis matrix, $G(q)\!\in\!\mathbb{R}^n$ is a moment term due to gravity, $T\!\in\!\mathbb{R}^n$ is a control input given in torque, and $T_e\!\in\!\mathbb{R}^n$ is an external disturbance in torque. 
\section{Selection of Distance Metric and Potential Function}\label{Sec:3}
In this section, we first describe how we can represent the difference of two points on the manifold $G$. Let $g_e, g_d \in G$, where $g_e$ is a current configuration of the end-effector, and $g_d$ is a desired configuration. We denote $g_e$ as $g$, i.e., drop the subscript if unnecessary. Then, the \emph{error} between two configurations $g$ and $g_d$ can be represented.
\begin{equation}
    g_{de} = g_d^{-1}g = \begin{bmatrix}
        R_d^T R & R_d^T(p - p_d)\\
        0 & 1
    \end{bmatrix},
\end{equation}
where $g = (R,p)$, $g_d = (R_d, p_d)$, and $g_{de} = (R_{de}, p_{de})$.
This error notation is widely utilized to define the configuration error in many of the geometric control literature, e.g., \cite{seo2023geometric, bullo1995proportional, lee2010geometric,bullo1999tracking,teng2022lie,lee2012exponential}. 

We interpret the potential function as the weighted version of the squared distance metric, as analogous to the potential function of the linear spring in Euclidean space. Then, the question of how to select the potential function now boils down to how to first represent the \emph{distance} between two points on the manifold. 
There are two major aspects of the $\SE$ group and manifold structure that one could leverage in order to define the distance:
\begin{enumerate}
    \item $\SE$ as a matrix group (as can be seen in homogeneous matrix representation)
    \item Property of Lie algebra (being a local chart of the manifold)
\end{enumerate}
The former aspect leads to the definition of distance and potential function on the Lie group \cite{seo2023geometric, bullo1999tracking, lee2010geometric}, and the latter one leads to the potential function on the Lie algebra \cite{teng2022lie,bullo1995proportional}. Here, we will refer to the squared distance metric as an error function.
\subsection{Error and Potential Function Definition on the Lie Group}
In \cite{lee2010geometric}, the error function $\Psi_{\SO}$ on $\SO$ between $R$ and $R_d$ is defined as follows:
\begin{equation}
    \Psi_{\SO}(R,R_d) = \tr(I - R_d^TR)
\end{equation}
Though this notation of error function was widely utilized in the Unmanned Aerial Vehicle (UAV) field after its suggestion in \cite{lee2010geometric}, its origin dates back to earlier works in \cite{koditschek1989application, bullo1999tracking}.
As shown in \cite{huynh2009metrics}, the error function $\Psi_{\SO}$ can be generalized utilizing the Frobenius norm. Utilizing the definition of the Frobenius norm, i.e., $\|A\|_F^2 = \tr(A^TA) = \tr(AA^T)$, and property of the trace operator that matrix (cyclically) commutes within the trace operator, it can be shown that
\begin{equation} \label{eq:error_fun_SO(3)}
    \begin{split}
        \Psi_{\SO}(R,R_d) &= \tfrac{1}{2}\tr\left((I - R_d^TR)^T(I - R_d^T R)\right) \\
        &= \tfrac{1}{2}\|I - R_d^T R\|_F^2
    \end{split}    
\end{equation}
The main takeaway here is that the distance between two configuration matrices can be identified via a Frobenius norm, which is a standard matrix norm. 
Noting that the transpose on $\SO$ is identical to inverse, $R^T = R^{-1}$, the error function on $\SO$ can be generalized to $\SE$ as follows \cite{seo2023geometric}:
\begin{equation} \label{eq:error_fun_SE(3)}
    \begin{split}
        & \Psi_{\SE}(g,g_d) = \tfrac{1}{2} \|I - g_d^{-1}g\|_F^2 \\
        &\;\; = \tfrac{1}{2} \tr\left((I - g_d^{-1}g)^T(I - g_d^{-1}g)^T \right) = \tfrac{1}{2} \tr (X^T X)\\
        &\;\; = \tr(I - R_d^T R) + \tfrac{1}{2}(p - p_d)^T(p - p_d)\\
        &\;\; = \Psi_{\SO}(R,R_d) + \tfrac{1}{2}\|p - p_d\|^2_2
    \end{split}
\end{equation}
with
\begin{equation*}
    X = X(g,g_d) = \begin{bmatrix}
        (I - R_d^TR) & -R_d^T(p - p_d) \\
        0 & 0 
    \end{bmatrix} 
\end{equation*}
For the rest of the paper, we drop the subscript $\SE$ and denote $\Psi_{\SE}(g,g_d)$ as $\Psi_1(g,g_d)$ for simplicity. A subscript $1$ is added to distinguish the Frobenius norm-based error function from the error function defined later in this section. The interpretation of the error function $\Psi_{\SO}$ as a Frobenius norm was not highlighted before \cite{seo2023geometric}, but it is a crucial key point to a direct extension of the error function from $\SO$ to $\SE$. 

A potential function can be obtained by properly weighting the distance metric. We utilize the weighted matrix $X_K$ as
\begin{equation}
    X_K = \begin{bmatrix}
        \sqrt{K_R}(I - R_d^TR) & -\sqrt{K_p}R_d^T(p - p_d)\\
        0 & 0
    \end{bmatrix}
\end{equation}
where $K_p, K_R \in \mathbb{R}^{3\times3}$ are symmetric positive definite stiffness matrices for translational and rotational parts, respectively, and the square root of the matrix can be defined from the singular value decomposition. Then, the potential function $P_1(g,g_d)$ on Lie group $\SE$ is defined in the following way:
\begin{align} \label{eq:potential_lie_group}
    & P_1(g,g_d) = \tfrac{1}{2}\tr(X_K^T X_K) \\
    &\;\;= \tr(K_R(I - R_d^T R)) + \tfrac{1}{2}(p - p_d)^T R_d K_p R_d^T (p - p_d)  \nonumber
\end{align}
We present a detailed description of the potential function \eqref{eq:potential_lie_group} in the following remark. 
\begin{remark} \textbf{Comments on the potential function \eqref{eq:potential_lie_group}}\\
1. The potential function $P_1(g,g_d)$ is left invariant, i.e., $P_1(g_lg,g_lg_d)$, where $g_l \in G \subset \SE$ is an arbitrary rigid body transformation acting on the left-hand side. It is also positive definite and is locally quadratic \cite{bullo1999tracking}. \\
2. By defining the error function from the Frobenius norm, we state the potential function $P_1(g,g_d)$ is a \emph{natural} potential function on $\SE$ among many candidate potentials in \cite{bullo1999tracking}, where no justification on the selection of the potential function was provided.\\
3. The potential function in \cite{lee2010geometric} is defined on $\SO$, and it only allows scalar gains. On the other hand, the rotational part of the potential functions in \cite{seo2023geometric, bullo1999tracking} allows matrix-based gains, which is crucial in the context of variable impedance control formulation in robotics applications.
\end{remark}

\subsection{Error and Potential Function Definition on the Lie Algebra}
In the previous section, the Lie algebra, of the group element $g  = (R,p) \in G$, $\hat{\xi} = \log_{\SE}(g)$ can be derived as
\begin{equation*}
        \hat{\xi} = \begin{bmatrix}
        \hat{\psi} & A^{-1}(\psi)p\\
        0 & 0 
    \end{bmatrix} \in \gfrak, \quad \xi = \begin{bmatrix}
        A^{-1}(\psi) p \\ \psi
    \end{bmatrix}\in \mathbb{R}^6.
\end{equation*}
As mentioned in \cite{bullo1995proportional}, the inner product on $\so$ is well defined and has desirable properties of positive definite and bi-invariant. The inner product of two elements $\hat{\psi}_1, \hat{\psi}_2 \in \so$ is defined using the Killing form, a bilinear operator defined by $\langle \cdot, \cdot \rangle _K: \so \times \so \to \mathbb{R}$, as
\begin{equation*}
    \begin{split}
        \langle\hat{\psi}_1, \hat{\psi}_2\rangle_{\so} &\triangleq -\tfrac{1}{4} \langle\hat{\psi}_1, \hat{\psi}_2\rangle_K, \;
        \langle\hat{\psi}_1, \hat{\psi}_2\rangle_K \triangleq \tr( \ad_{\psi_1} \ad_{\psi_2} ).
    \end{split}    
\end{equation*}

In contrast, there is a negative result on $\se$: there is no symmetric bilinear form on $\se$ that is both positive definite and bi-invariant \cite{bullo1995proportional, park1995distance}.
Still, one can define a left-invariant inner product on $\se$ by exploiting the fact that $\se$ is isomorphic to $\mathbb{R}^6$. We will denote the inner product by $\langle \hat{\xi}_1, \hat{\xi_2}\rangle_{(P,I)}$, as follows \cite{park1995distance, teng2022lie, vzefran1996choice, belta2002euclidean}:
\begin{equation}
\label{eq:inner_product_se(3)}
    \langle \hat{\xi}_1, \hat{\xi_2} \rangle_{(P,I)} = \xi_1^T P \xi_2
\end{equation}
where $\hat{\xi}_1, \hat{\xi}_2 \in \gfrak$, and $P \in \mathbb{R}^{6\times 6}$ is symmetric positive definite matrix. 
The subscript $P$ in definition \eqref{eq:inner_product_se(3)} indicates that the inner product is equipped with the Riemannian metric $P$, while the subscript $I$ represents the fact that the Lie algebra $\se$ is a  vector space tangent to the identity of $\SE$. 
Then, one can define the Riemannian metric on $\SE$ on the point $g \in G$,
using the inner product on $\se$ $\langle \cdot, \cdot \rangle_{(P,g)} : T_gG\times T_gG \to \mathbb{R}$, as follows. Consider two tangent \emph{vectors} (perturbation on the configuration manifold) on the tangent space at point g, $\delta g_i \in T_gG$, such that $\delta g_i = g \hat{\xi}_i$ with $\hat{\xi}_i = (\hat{\psi}_i, b_i)$ for $i = \{1,2\}$.
The Riemannian metric is then defined as \cite{park1995distance, vzefran1996choice, belta2002euclidean}
\begin{equation} \label{eq:Riemannian_metric}
    \langle \delta g_1, \delta g_2 \rangle_{(P,g)} \!=\! \langle g^{-1} \delta g_1, g^{-1} \delta g_2 \rangle_{(P,I)} = \langle \hat{\xi}_1, \hat{\xi}_2 \rangle_{(P,I)}
\end{equation}
The Riemannian metric defined in \eqref{eq:Riemannian_metric} is often referred to as a left-invariant Riemannian metric since,  $g^{-1} \delta g_i = \hat{\xi}_i$ is left invariant, i.e., for any arbitrary left transformation matrix $g_l$, $(g_l g)^{-1} (g_l g) \hat{\xi}_i = \hat{\xi}_i$. 
By letting $P = \tfrac{1}{2}I_{6\times6}$, we can define an error function as
\begin{equation} \label{eq:error_function_SE3_algebra}
    \Psi_2(g,g_d) = \langle g_{de}\hat{\xi}_{de}, g_{de}\hat{\xi}_{de} \rangle_{\left(0.5I, g_{de}\right)} = \tfrac{1}{2} \|\psi_{de}\|^2 + \tfrac{1}{2}\|b_{de}\|^2
\end{equation}
where $\hat{\xi}_{de} = \log(g_{de}) = (\hat{\psi}_{de}, b_{de})$. 
Eq. \eqref{eq:error_function_SE3_algebra} can also be obtained  by considering  $\SE$ as a general  affine group, as  done in \cite{belta2002euclidean}  and \cite{liu2013finite}. 
However as noted  \cite{park1995distance},  the error function $\Psi_2(g,g_d)$ is not  a (squared) geodesic distance. 
The geodesic distance proposed in \cite{park1995distance} is given by:
\begin{equation*}
    \Psi_{geo,\SE} = \tfrac{1}{2}\|\psi_{de}\|^2 + \tfrac{1}{2}\|p - p_d\|^2,
\end{equation*}
but since $\text{det}(A^{-1}(\psi)) \neq 1$, $\|p - p_d\| \neq \|b_{de}\|$. 

Given the inner product on $\se$ \eqref{eq:inner_product_se(3)}, the potential function can now be defined as  \cite{teng2022lie}
\begin{equation} \label{eq:potential_lie_algebra}
    P_2(g,g_d) = \langle \hat{\xi}_{de}, \hat{\xi}_{de}\rangle_{(0.5K_\xi, I)} = \tfrac{1}{2}\xi_{de}^T K_{\xi} \xi_{de},
\end{equation}
where $P=\tfrac{1}{2}K_{\xi}$ and $K_\xi \in \mathbb{R}^{6\times6}$ is symmetric positive definite. Henceforth, we interpret $K_\xi$ as a stiffness matrix. Notice that 
$P_2(g,g_d)$ is left-invariant, positive definite, and quadratic.

\section{Control Design and Stability Result}
\subsection{Velocity Error Definition}
Interestingly, both approaches utilize the same velocity error given as follows:
\begin{equation} \label{eq:velocity_error}
    e_V = V^b - V^*_{d}, \quad \text{with} \;  V^*_{d} =  \Ad_{g_{ed}}V_d^b
\end{equation}
where $\Ad_{g_{ed}} = \Ad_{g_{de}^{-1}}$. However, the underlying idea of how the velocity error is derived differs slightly. 

In the Frobenius norm-based approach, the tangent vectors $\dot{g} \in T_gG$ and $\dot{g}_d \in T_{g_d}G$ are first compared. Since the tangent vectors lying on the different tangent space cannot be directly compared, the desired tangent vector $\dot{g}_d$ is first transferred to the tangent space of $g$, $T_gG$ by multiplying $g_{de}$ to the righthand side of $\dot{g}_d$ \cite{lee2010geometric, seo2023geometric} as follows:
\begin{align*}
    \dot{g} \!-\! \dot{g}_d(g_d^{-1}g) \!&=\! g \hat{V}^b \!-\! g_d \hat{V}^b_d g_d^{-1}g \!=\! g\left(\hat{V}^b \!-\! g^{-1}g_d \hat{V}^b_d g_d^{-1}g\right) \nonumber \\
    &= g\hat{e}_V \in T_{g}G
\end{align*}

On the other hand, the configuration error $g_{de}$ is directly interpreted as an element on the manifold $G$, and the time-derivative of $g_{de} = g_d^{-1}g$ is dealt with in \cite{teng2022lie} in the following way.
\begin{align*}
    \dfrac{d}{dt}(g_d^{-1}g) &=  g_d^{-1}\dot{g} + \dot{g}_d^{-1} g = g_d^{-1}\dot{g} +  g_d^{-1}\dot{g}_d g_d^{-1} g \nonumber \\
    &= g_d^{-1}g \hat{V}^b + g_d^{-1}g_d\hat{V}_d^b g_d^{-1}g \\
    &= g_d^{-1}g \left(\hat{V}^b + (g_d^{-1}g)^{-1}\hat{V}_d^b (g_d^{-1}g)\right) \triangleq g_d^{-1}g (\hat{e}_V) \nonumber\\
    \implies \dot{g}_{de} &= g_{de} \hat{e}_V \in T_{g_{de}}G\nonumber
\end{align*}
Recalling that $g_{de} = g_d^{-1}g$, $g_{de}^{-1} = g_{ed}$, and $(g \hat{\xi} g^{-1})^\vee = \Ad_g\xi$, for $g \in G$ and $\xi \in \gfrak$, we show that the velocity error for both approaches are identical.

\subsection{Impedance Control as Dissipative Control}
As suggested in \cite{seo2023geometric}, we consider the impedance control as a dissipative control design. Consider a Lyapunov function similar to the total mechanical energy of the system
\begin{equation} \label{eq:lyapunov}
    V_i(t,q,\dot{q}) = K(t,q,\dot{q}) + P_i(t,q), \quad i \in \{1,2\}
\end{equation}
where $K$ and $P_i$ are the kinetic and potential energy components of the Lyapunov function. Note that $P_1$ is given in \eqref{eq:potential_lie_group} and $P_2$ is in \eqref{eq:potential_lie_algebra}. Note also that we have changed the argument of $P$ from $(g,g_d)$ to $(t,q)$ for the compactness of notation. This is because $g_d$ only depends on time $t$, and $g$ is a function of the generalized coordinate vector $q$, connected by forward kinematics. The kinetic energy component of the Lyapunov function is defined as follows:
\begin{equation} \label{eq:kinetic_energy}
    K(t,q,\dot{q}) = \tfrac{1}{2}e_V^T \tilde{M} e_V,
\end{equation}
The desired property of the impedance control to be dissipative \cite{kronander2016stability} is defined as
\begin{equation} \label{eq:dissipative}
    \dot{V}_i = -e_V^T K_d e_V
\end{equation}
where $K_d \in \mathbb{R}^6$ is symmetric positive definite damping matrix. 
To satisfy the dissipative property \eqref{eq:dissipative}, we consider the following control laws
\begin{equation} \label{eq:control_law}
        \tilde{T}_i \!=\! \tilde{M}\dot{V}_{^*d} \!+\! \tilde{C}{V}_{^*d} + \tilde{G} \!-\! f_{_{G,i}} \!-\! K_d e_V, \;\text{for}\; i \in \{1,2\}
\end{equation}
where $f_{_{G,i}} \in \mathbb{R}^6$, $\hat{f}_{_{G,i}}\in \se^*$, are given by either
\begin{equation}
    \begin{split}
        f_{_{G,1}} &= \begin{bmatrix}
        f_p \\ f_R
        \end{bmatrix} = \begin{bmatrix}
            R^T R_d K_p R_d^T (p - p_d)\\
            (K_R R_d^T R - R^T R_d K_R)^\vee
        \end{bmatrix} \;\; \text{or} \\ 
        f_{_{G,2}} &= K_\xi \xi_{de}
    \end{split}    
\end{equation}
For the rest of the paper, we will name the $\tilde{T}_1$ control law as $\CLG$ and the $\tilde{T}_2$ control law as $\CLA$.
The following theorem shows that the control law \eqref{eq:control_law} satisfies the dissipative property of the Lyapunov function.
Note that the control laws are based on the following assumption.
\begin{assumption}\label{assum:1}
    The end-effector lies in a region $\left.D \subset \SE\right.$ such that the Jacobian $J_b$ is full-rank, i.e., non-singular. Moreover, the end-effector of the manipulator and the desired trajectory lies in the reachable set $\mathcal{R}$, i.e., $p(\theta)\!\in\!\mathcal{R}\!=\!\{ p(\theta) \;|\; \forall\theta\!\in\!\mathbb{S}^n\}\!\subset\!\mathbb{R}^3$. 
    The desired trajectory is also continuously differentiable.
\end{assumption}
\begin{theorem}
    Suppose assumption \ref{assum:1} holds true and there are no external disturbances, i.e., $T_e = 0$. Consider a robotic manipulator with dynamics \eqref{eq:robot_dynamics_eef} and energy-based Lyapunov function candidate \eqref{eq:lyapunov}. Then, the Lyapunov function of the closed-loop system with the control law \eqref{eq:control_law} satisfies the dissipative property \eqref{eq:dissipative}, for $i = 1,2$, respectively.
\end{theorem}
\begin{proof}
    For the case when $i=1$, the result is already shown in \cite{seo2023geometric}.
    For the case when $i=2$, the proof can be done similarly in \cite{seo2023geometric}, and it suffices to show that,
    \begin{equation} \label{eq:potential_algebra_derivative}
        \begin{split}
            \dot{P}_2(t,q) &= \langle \hat{\xi}_{de}, \hat{e}_V \rangle_{(K_\xi,I)} = \xi_{de}^T K_\xi e_V =  f_{_{G,2}}^T e_V,
        \end{split}
    \end{equation}
    where \eqref{eq:potential_algebra_derivative} are shown in \cite{teng2022lie} -- See Eq (14) and Theorem 1 in \cite{teng2022lie}. 
\end{proof}
The stability results of the controllers are presented in the following theorem.
\begin{theorem}
    Suppose assumption~\ref{assum:1} holds true and there are no external disturbances, i.e., $T_e = 0$. Consider a robotic manipulator with dynamics \eqref{eq:robot_dynamics_eef} and energy-based Lyapunov function candidate \eqref{eq:lyapunov}.
    The closed-loop system with control law \eqref{eq:control_law} is asymptotically stable. 
\end{theorem}
\begin{proof}
    For the case $i=1$, the result for asymptotic stability is shown in \cite{seo2023geometric}. For the case $i=2$, the proof for the stability is similar to \cite{seo2023geometric} with the introduction of a coupling term \cite{bullo1995proportional,bullo1999tracking,lee2010geometric} given by 
    \begin{equation*}
        V_2' = V_2 + \langle \hat{\xi}_{de}, \hat{e}_V \rangle_{(\varepsilon I, I)} = V_2 + \varepsilon \xi_{de}^T e_V, \quad \varepsilon > 0.
    \end{equation*}
    The relation between the time-derivative $\dot{\xi}_{de}$ and $e_V$ should be identified to analyze the time-derivative of the coupling term, The relation is presented in \cite{bullo1995proportional}, and is as follows:
    \begin{equation*}
        \dot{\xi}_{de} = \mathcal{B}_{\xi_{de}} e_V, \quad \text{where}\;\; \mathcal{B}_{\xi} \triangleq \sum_{n=0}^{\infty}  \dfrac{(-1)^n B_n}{n!}(\ad_{\xi})^n
    \end{equation*}
    where $\ad_\xi$ is provided in \eqref{eq:Ad_and_ad}, and $B_n$ is Bernoulli number, e.g., $B_0 = 1$, $B_1 = -1/2$, $B_2 = 1/6$, $B_3 = 0$, $B_4 = -1/30$ and so forth. Then, the rest of the proof can be done following Theorem 6 in \cite{bullo1995proportional}.
\end{proof}
The properties of the controllers and the detailed comparison results are provided in Section.~\ref{Sec:5}

\section{Comparison Analysis}\label{Sec:5}
\subsection{Qualitative Comparison}
In this subsection, we will compare the properties of the control laws utilizing different potential functions: Lie group-based and Lie algebra-based. For the rest of the paper, we consider $K_\xi$ in $\fgtwo$ to be
$K_\xi = \text{blkdiag}(K_b, K_{\psi})$ to enable a fair comparison with $\fgone$, where the gains for translational and rotational dynamics are separated. Here, $K_b, K_{\psi} \in \mathbb{R}^{3\times 3}$ stands for symmetric positive definite translational and rotational stiffness matrices. 
\subsubsection{Regarding Potential Function}
In the case of $K_\xi = \text{blkdiag}(K_b, K_{\psi})$, the potential function $P_2(g,g_d)$ becomes
\begin{align} \label{eq:potential_lie_algebra_separated}
    P_2(g,g_d) &= \tfrac{1}{2}\psi_{de}^T K_\psi \psi_{de}\\ &+
    \tfrac{1}{2}(p - p_d)^TR_dA^{-T}(\psi) K_b  A^{-1}(\psi) R_d^T (p-p_d) \nonumber
\end{align}
From \eqref{eq:potential_lie_group}, one can easily notice that the potential function $P_1(g,g_d)$ is separated into its translational and rotational part. Noting that as $\det(R_d) = 1$ from the definition of the rotational matrix, the size of the translational potential function is the same as the potential function of the Euclidean case, i.e.,
\begin{equation*}
    \tfrac{1}{2}(p-p_d)^TK_p (p - p_d) = \tfrac{1}{2}(p - p_d)^T R_d K_p R_d^T (p - p_d).
\end{equation*}
Therefore, the rotational error does not affect the translational potential function in $P_1(g,g_d)$. On the other hand, comparing $\eqref{eq:potential_lie_algebra_separated}$ with \eqref{eq:potential_lie_group},
\begin{align*}
    &\tfrac{1}{2}(p-p_d)^TK_p (p - p_d) \\
    &\neq \tfrac{1}{2}(p - p_d)^TR_dA^{-T}(\psi) K_b  A^{-1}(\psi) R_d^T (p-p_d) 
\end{align*}
From the fact that $\text{det}(A^{-1}(\psi))\neq 1$. Therefore, the rotational error does affect the translational potential function in $P_2(g,g_d)$ and the translational P control actions $\fgtwo$ as well.

As a side note, the unbounded potential function $\SO$ was proposed in \cite{jeong2022memory}. Noting that the Lie algebra-based potential function of $\SE$ is written as a summation of the translational and the rotational potential function, the unbounded potential function on $\SO$ could be incorporated into the $\SE$ potential function. 
\subsubsection{Regarding Positional Error}
After separating the gains, $\fgtwo$ now becomes
\begin{equation}
    f_{_{G,2}} = \begin{bmatrix}
        K_b b_{de}\\
        K_\psi \psi_{de}
    \end{bmatrix} = \begin{bmatrix}
        K_b A^{-1}(\psi) R_d^T(p - p_d) \\
        K_\psi \log_{\SO}(R_d^T R)
    \end{bmatrix},
\end{equation}
Both controllers' elastic forces $\fgone, \fgtwo$ share the common property that rotational error affects the translational control inputs. This property shows a clear difference between geometric impedance control and Cartesian impedance control, where the translational and the rotational errors are handled separately. 

However, as described in the previous part, the $\fgtwo$ differs from $\fgone$, in that the rotational error affects the magnitude of the translational elastic force. Therefore, there might be a case that even though when the translational error is small, the magnitude of translational elastic force is not small if the rotational error is large. Although from the formulation of $A^{-1}(\psi)$, and from the formulation of the rotational matrix, the magnitude of $A^{-1}(\psi)$ is bounded and is not significantly big. 

It is worth noting that the positional error vector $\log{(g_{de})}$ does not lie in the gradient direction of geodesics. In \eqref{eq:potential_algebra_derivative}, it is shown that $\fgtwo$ (with $K_\psi = I$) lies on the gradient direction of error function $\Psi_2(g,g_d)$ \eqref{eq:error_function_SE3_algebra}, and we showed that $\Psi_2(g,g_d) \neq \Psi_{geo,\SE}(g,g_d)$.
\subsubsection{Regarding Velocity Error}
\begin{figure}
    \centering
    \begin{subfigure}{0.49\columnwidth}
        \centering
        \includegraphics[width=1\linewidth]{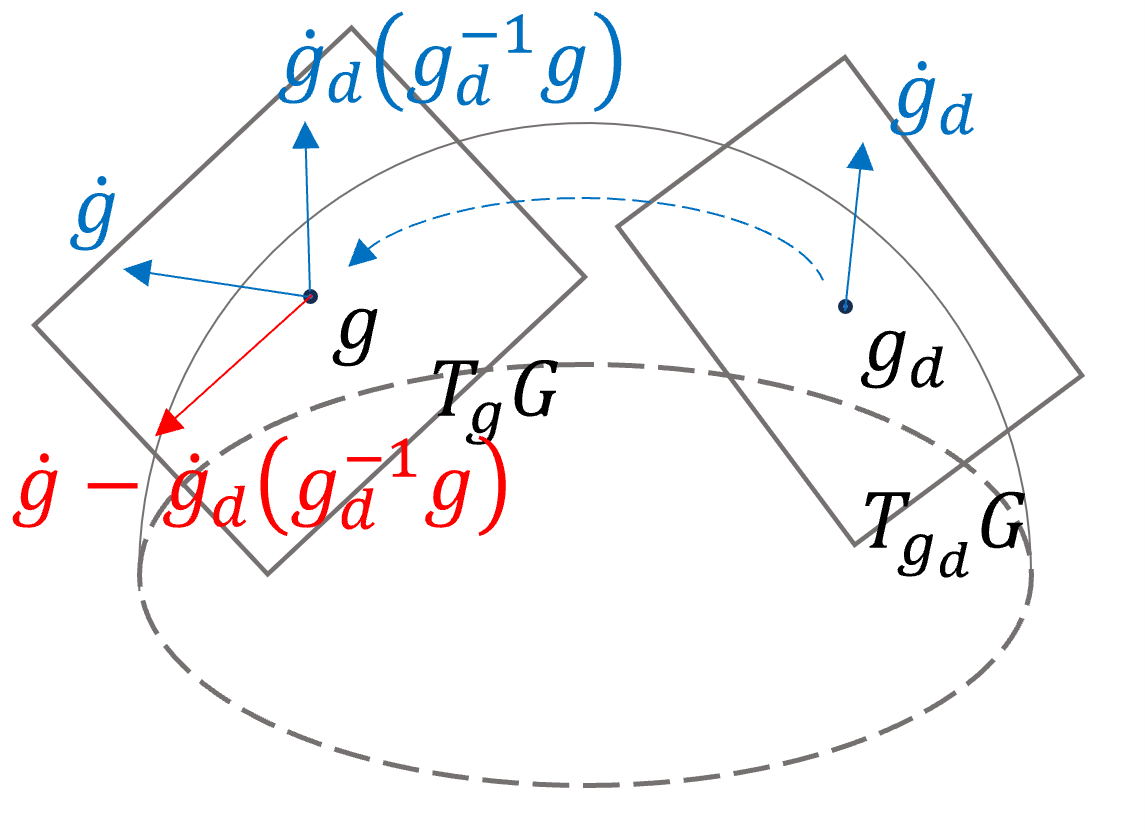}
        \caption{}
    \end{subfigure}
    \begin{subfigure}{0.49\columnwidth}
        \centering
        \includegraphics[width=0.90\linewidth]{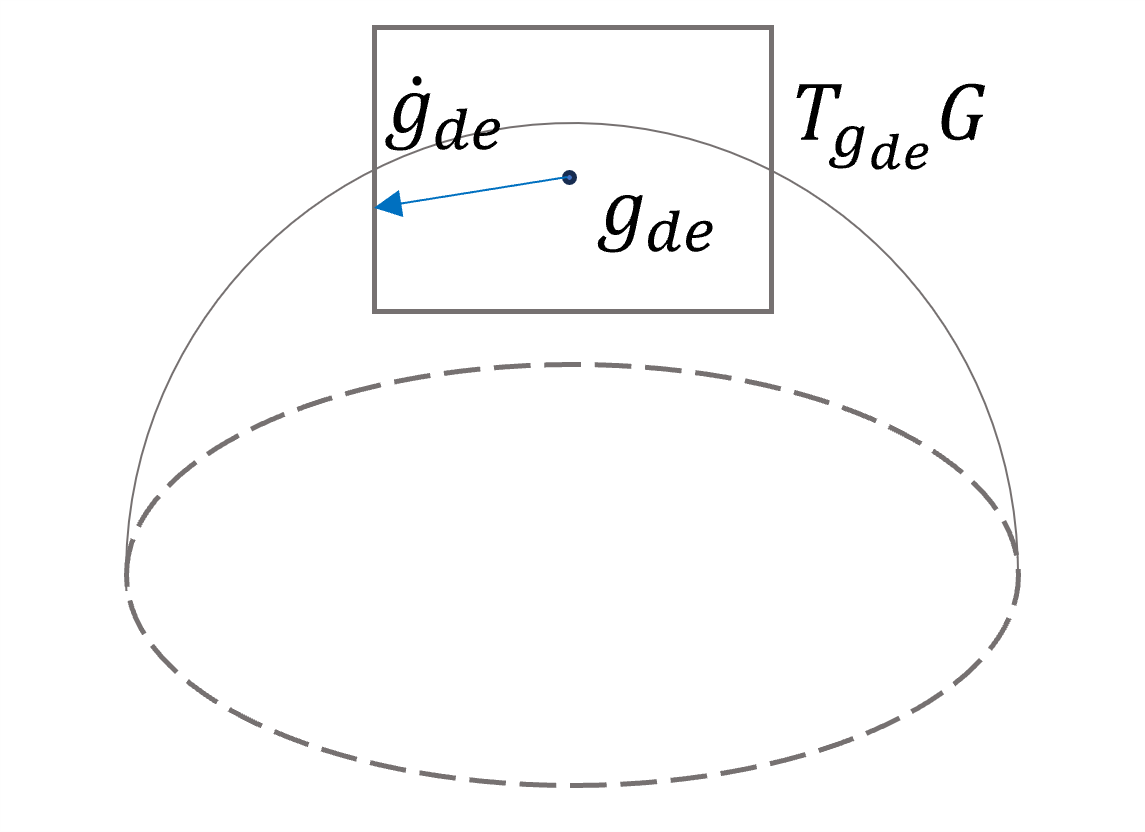}
        \caption{}
    \end{subfigure}
    \caption{The concept behind the definition of the velocity error. (a) $\dot{g}_d \in T_{g_d}G$ is first translated via right multiplication $g_d^{-1}g$. (b) $g_{de}$ as a whole is interpreted as a group element and the vector $\dot{g}_{de} \in T_{g_{de}}G$ is analyzed directly.}
    \label{fig:comparison_velocity_error}
\end{figure}

Although both controllers utilize the same velocity error \eqref{eq:velocity_error}, the rationale behind the two approaches is quite different -- See Fig.~\ref{fig:comparison_velocity_error}. In Lie group-based derivation, the two tangent vectors $\dot{g} \in T_gG$ and $\dot{g}_d \in T_{g_d}G$ cannot be directly compared since they are lying on the different tangent spaces. Therefore, to compare the tangent vectors on the tangent space $T_gG$, $\dot{g}_d$ is first translated via right multiplication $g_d^{-1}g$. Note that this right multiplication action can be considered as pushforward mapping since the action is translating a tangent vector from one tangent space to another tangent space. 

On the other hand, in Lie algebra-based formulation, the configuration error $g_{de} = g_d^{-1}g$ is directly interpreted as a component on $G \subset \SE$. The tangent vector is then directly derived on the tangent space $\dot{g}_{de} \in T_{g_{de}}G$, and it turns out that $\dot{g}_{de} = g_{de} \hat{e}_V$, leading to same velocity error formulation. 

The definition of the velocity error enables the definition of the kinetic energy \eqref{eq:kinetic_energy} and dissipated energy \eqref{eq:dissipative}, using the Riemannian metric \eqref{eq:Riemannian_metric}. One interesting point is that the Riemannian metrics are defined on different points depending on the perspectives. For the Lie group-based design, the Kinetic energy $K$ and dissipative energy term $D$ can be defined as
\begin{equation*}
    \begin{split}
        K &= \langle \dot{g} - \dot{g}_d(g_d^{-1}g),\dot{g} - \dot{g}_d(g_d^{-1}g) \rangle_{(0.5\tilde{M},g)} \\ &= \langle \hat{e}_V, \hat{e}_V \rangle_{(0.5\tilde{M},I)} = \tfrac{1}{2}e_V^T \tilde{M} e_V,\\
        D &= \langle \dot{g} - \dot{g}_d(g_d^{-1}g),\dot{g} - \dot{g}_d(g_d^{-1}g) \rangle_{(K_d,g)} = e_V^T K_d e_V.
    \end{split}
\end{equation*}
On the other hand, in the Lie algebra-based design, $K$ and $D$ terms are similarly defined as
\begin{equation*}
    \begin{split}
        K&= \langle \dot{g}_{de}, \dot{g}_{de} \rangle_{(0.5\tilde{M},g_{de})} = \tfrac{1}{2}e_V^T \tilde{M} e_V\\
        D&= \langle \dot{g}_{de}, \dot{g}_{de} \rangle_{(K_d,g_{de})} = e_V^T K_d e_V
    \end{split}
\end{equation*}
Therefore, the coincidence in the velocity error leads to the same GIC law formulation.

\subsubsection{Drawbacks of utilizing potential functions on the Lie group}
A well-known drawback of the Lie group-based potential function is that the rotational part $\tr(K_R (I - R_d^T R))$ is not quadratic in its domain. Consider the error function on $SO(2)$, $\Psi_{SO(2)}$ as an example. The rotation matrix on $SO(2)$ with some rotation angle $\theta$ is given by
\begin{equation*}
    R(\theta) = \begin{bmatrix}
        \cos{\theta} & -\sin{\theta}\\
        \sin{\theta} & \cos{\theta}\\
    \end{bmatrix}
\end{equation*}
Note that this is the case where the rotation matrix on $\SO$ is represented via axis-angle representation with the axis of rotation $\omega = [0,0,1]^T$ and the rotation angle around the axis is $\theta$. 
The configuration error between $R(\theta)$ and the identity $I = R(0)$ is $R(\theta) = I^{-1}R(\theta)$. Then, with the gains $K_R, K_\psi = I$, the P actions $f_{_{G,i}}$ on $SO(2)$ for both controllers are given as
\begin{equation*}
    \begin{split}
        f_{_{G,1}} \!=\! (R \!-\! R^T)^\vee  \!=\! 2 \sin(\theta), \;\; f_{_{G,2}} \!=\! \log(R)^\vee \!=\! \theta
    \end{split}
\end{equation*}
where $(\cdot)^\vee : \mathfrak{so}(2) \to \mathbb{R}$ is a vee map on $SO(2)$. Moreover, the error function $\Psi_1(R,I)$ and $\Psi_2(R,I)$ are
\begin{equation*}
    \Psi_1(R,I) = \tfrac{1}{2}\|I - R\|_F, \quad \Psi_2(R,I) = \tfrac{1}{2}\|\log(R)\|^2
\end{equation*}
\begin{figure}[t!]
    \hspace{-6pt}
    \input{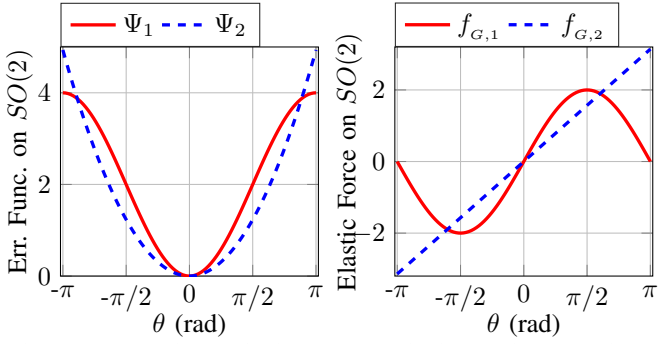}
    \vspace{-20pt}
    \caption{Comparison of Lie group and Lie algebra-based approaches for (Left) Error functions $\Psi_1(R,I)$ and $\Psi_2(R,I)$, (Right) Elastic forces $\fgone(R,I)$ and $\fgtwo(R,I)$ with identity gains.}
    \label{fig:comparison_SO2}
    \vspace{-7pt}
\end{figure}
Then, the comparison results on $SO(2)$ example are presented in Fig.~\ref{fig:comparison_SO2}. As can be seen in Fig.~\ref{fig:comparison_SO2}, the size of the P control action $\fgone$ diminishes to $0$ when the rotation matrix is completely flipped $180^\circ$. In addition, the error function of $\Psi_2$ is quadratic to the rotation angle, but $\Psi_1$ is only locally quadratic, i.e., $\Psi_1 = 2 (1 - \cos(\theta))$. Note that the trends on $SO(2)$ are extended to $\SO$ without loss of generality the axis of rotation $\omega$ can be chosen arbitrarily.

The geometric control approach on $\SO$ \cite{lee2010geometric} is widely utilized in the Unmanned Aerial Vehicle (UAV) field. However, the main criticism of this controller is that as the size of the rotational error becomes larger, the convergence speed of the closed-loop rotational dynamics tends to be slower. 
The convergence properties of the inner-loop rotational dynamics are of critical importance in this context. This is because geometric control for UAVs relies on a time-scale separation approach, wherein the control system is structured with slower outer-loop translational dynamics and faster inner-loop rotational dynamics. Consequently, if the fast convergence of the rotational dynamics is not guaranteed, the overall stability of the control system can be hampered. 

In \cite{lee2012exponential}, a potential function is re-designed to have a quadratic shape on an almost global region to overcome this drawback. However, the potential function proposed herein is somewhat artificial, i.e., it is not derived from natural norms, which can reduce the appeal of geometric control approaches. 
\subsubsection{Numerical property} 
Unlike the $SO(2)$ case where the log map of the rotation matrix is calculated as the angle, the log map of $\SO$ requires solving \eqref{eq:log_mapping_SO}. Therefore, the log map on $\SO$ needs division over $\sin{\phi}$, where $\phi$ is an angle difference in axis-angle representation. When the angle difference is small, it may lead to trivial numerical instability, whereas there is no numerical issue in calculating $\fgone$. Numerical calculation-wise, although there is no significant difference, $\CLG$ is more computationally favorable than $\CLA$.
\subsubsection{Stability property}
$P_1(g,g_d)$ is locally quadratic and $P_2(g,g_d)$ is almost globally quadratic as could be partially observed in Fig.~\ref{fig:comparison_SO2}. In fact, the exponential stability result of both controllers \eqref{eq:control_law} could be driven straightforwardly following the result of \cite{bullo1995proportional, bullo1999tracking}. However, one can only show the local exponential stability for the $\CLG$, while almost global exponential stability can be shown for the $\CLA$. 
\subsubsection{Equivariance property}
A key property of the geometric impedance control is that it can serve as an equivariant controller when described in the spatial frame \cite{seo2023contact}. As proposed in \cite{seo2023contact}, the PD control part of the controller is equivariant if it satisfies the following conditions: 1. if it is represented in the body frame and 2. if it is left-invariant in the body frame. The PD control action of $\CLG$ is shown to be equivariant in \cite{seo2023contact}. $\CLA$ in \eqref{eq:control_law} is represented in the body-frame coordinate by its design. 
Elastic force $\fgtwo(g,g_d)$ is left invariant as
\begin{align*}
    \fgtwo(g_lg,g_lg_d) &= K_\xi \log((g_lg_d)^{-1}(g_lg)) \nonumber\\
    &= K_\xi \log(g_d^{-1}g) = \fgtwo(g,g_d),
\end{align*}    
D control action $K_d e_V$ is left invariant as shown in \cite{seo2023contact} since both use the same D control input. This property shows a direct extension of the $\CLA$ in the framework of learning variable impedance control, where the impedance gains $K_\xi$ is learned by the neural network in a left-invariant manner with the input given by geometrically consistent error vectors (GCEV) \cite{seo2023contact}. Note that $\log(g_{de})$ can also serve as GCEV because of left-invariance. 

\subsection{Quantitative Comparison}
In this subsection, we present a quantitative comparison result of GICs via a numerical simulation. The controllers are tested on two cases: regulation with large initial error and trajectory tracking of a smooth trajectory. The simulation is conducted on Matlab, with a UR5e 6DOF robotic manipulator. Both controllers are equipped with the same gains. However, as shown in Fig.~\ref{fig:comparison_SO2}, the elastic forces $\fg$ have different magnitudes even with the same gains. The gains we utilize are
\begin{equation*}
    K_p \!=\! K_b \!=\! K_R \!=\! K_\psi \!=\! \text{diag}([100,100,100]), \;\; K_d = k_d I,
\end{equation*}
where $k_d = 50$. 
\subsubsection{Regulation with large initial error}
\begin{figure}[t!]
    \vspace{-5pt}
    \hspace{-8pt}
    \input{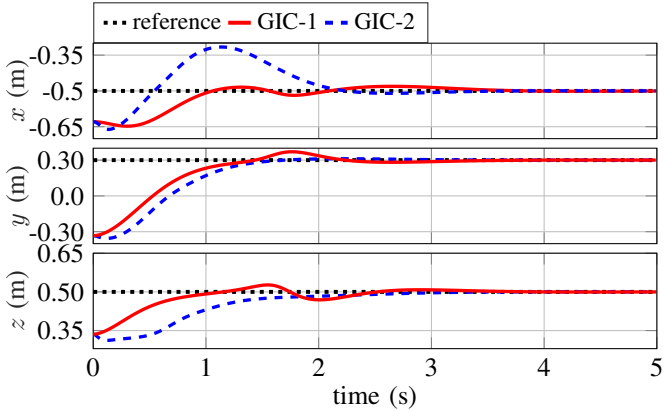}
    \vspace{-20pt}
    \caption{The regulation results with large initial errors in $x$, $y$, and $z$ coordinates are plotted for $\CLG$ in red solid lines and $\CLA$ in blue dashed lines.}
    \label{fig:comparison_xyz_regulation}
\end{figure}

In the numerical simulation, the initial rotational error is selected as $\pi-\epsilon$ in the axis-angle rotation, where $0<\epsilon \ll1$ is a small number. The resulting trajectories in $x$, $y$, and $z$ coordinates are presented in Fig.~\ref{fig:comparison_xyz_regulation}. We utilize the error function \eqref{eq:error_fun_SE(3)} to evaluate the convergence performance of the controllers, and the results are shown in Fig.~\ref{fig:comparison_xyz_regulation}. The performance is also quantitatively evaluated using the Root-Mean-Square (RMS) value of the errors and error functions in Table.~\ref{table:RMS_regulation}. 
As predicted in the qualitative comparison, the Lie group-based control formulation $\CLG$ showed poor convergence performance when the initial rotational error was big, but the Lie algebra-based control formulation $\CLA$ showed fair convergence. 
\begin{figure}
    \vspace{-5pt}
    \hspace{-6pt}
     \input{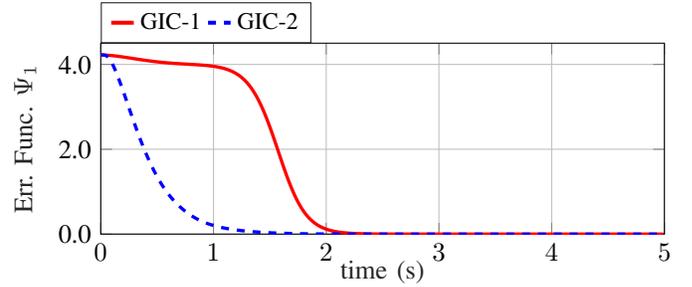}
    \caption{Error function  \eqref{eq:error_fun_SE(3)} $\Psi_1(g,g_d)$ for the resulting trajectories.}
    \label{fig:err_fun_regulation}
\end{figure}

\begin{table}
    \setlength\doublerulesep{0.5pt}
    \renewcommand\tabularxcolumn[1]{m{#1}}
    \centering
    \caption{Root-Mean-Squared value of errors for the regulation with large initial errors.}
    \label{table:RMS_regulation}
    \begin{tabularx}{\linewidth}{
     >{\centering\arraybackslash\hsize=1.4\hsize}X >{\centering\arraybackslash \hsize=0.8\hsize}X >{\centering\arraybackslash \hsize=0.8\hsize}X 
    }
    \toprule[1pt]\midrule[0.3pt]
     RMS values & $\CLG$ & $\CLA$ \\
    \midrule
    $\text{RMS}(x - x_d)$ & 0.0596 & 0.0784\\
    $\text{RMS}(y - y_d)$ & 0.1764 & 0.2104\\
    $\text{RMS}(z- z_d)$ & 0.0409 & 0.0704\\
    \midrule
    $\text{RMS}(\Psi_{\SO})$ \eqref{eq:error_fun_SO(3)} & 2.1519 & 0.9539\\
    $\text{RMS}(\Psi_{\SE})$ \eqref{eq:error_fun_SE(3)} & 2.1846 & 1.0206\\
    \midrule[0.3pt]\bottomrule[1pt]
    \end{tabularx}
\end{table}

\subsubsection{Trajectory tracking of a smooth trajectory}
\begin{figure}[t!]
    \vspace{-5pt}
    \hspace{-8pt}
    \input{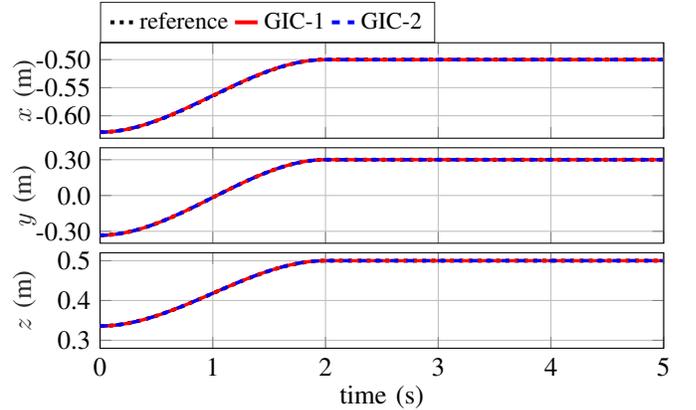}
    \vspace{-20pt}
    \caption{The trajectory tracking results with smooth desired trajectories in $x$, $y$, and $z$ coordinates are plotted for $\CLG$ in red solid lines and $\CLA$ in blue dashed lines. All lines are on top of each other and indistinguishable.}
    \label{fig:comparison_xyz_tracking}
\end{figure}

\begin{figure}
    \vspace{-5pt}
    \hspace{-8pt}
    \input{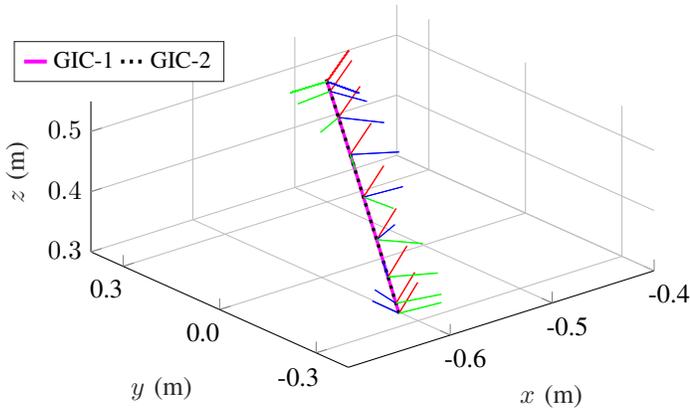}
    \caption{Trajectory results for $t = 0 \sim 5 (\mathrm{s})$ with end-effector axes attached are plotted. The trajectory results of both controllers are on top of each other and cannot be distinguished.}
    \label{fig:3d_traj}
\end{figure}

Using the same initial condition and goal position from the previous regulation case, here we design a smooth trajectory and test the trajectory tracking result. The trajectory is generated as the $3\textsuperscript{rd}$ polynomial order of time. The resulting trajectories in $x$, $y$, and $z$ coordinates are presented in Fig.~\ref{fig:comparison_xyz_tracking}, and 3D trajectory results are shown in Fig.~\ref{fig:3d_traj}. As can be seen in the figure, both controllers showed perfect trajectory tracking performance to smooth trajectories, and the convergence speed could be designed arbitrarily fast.  
We finalize the quantitative comparison with the remark explaining the meaning of the second simulation.
\begin{remark}
    The effect of main criticism for the Lie group-based control design $\CLG$ was shown in the numerical analysis with large initial errors. Although the drawback is critical in underactuated UAV systems, it is not necessarily a significant issue in the manipulator, where the system is normally fully actuated (6-DOFs) and frequently over-actuated (7-DOFs). This is because, for the fully actuated system, the consequence of the slow convergence does not lead to the destabilization of the whole control system. 
    Moreover, it is comparably easy for the manipulator system to get over large initial errors in heuristic ways, e.g., just slightly rotating the end-effector would solve the problem. 
    The second numerical simulation shows that one can easily deal with slow convergence by planning via smooth trajectories.
\end{remark}

\section{Conclusion}
In this paper, we provide a study to the question of how to design the potential function for geometric impedance control (GIC) on $\SE$. The two potential functions on $\SE$ are considered: a Lie group-based formulation and a Lie algebra-based formulation. 
The derivations of the distance metrics and the potential functions were presented in the context of differential geometry. 
Based on the potential function, the geometric impedance control laws were derived by considering impedance control as a dissipative control design.
Extensive qualitative and quantitative analyses were presented by comparing GIC laws derived from different potential functions. 
As a result, we showed that both GIC laws are legitimate control designs properly incorporating the geometric structure of the manipulator. 
We emphasize that both controllers contain the equivariance property in $\SE$, which is desirable in the context of learning manipulation tasks, possibly leading to learning transferability. 
A Lie algebra-based control design showed more favorable stability and convergence properties, while a Lie group-based control design is more mathematically concise and physically intuitive. 
Finally, we emphasized that the slow convergence of Lie group-based control may not be a critical issue in the manipulator system, as the issue can be easily handled via heuristics.
\bibliographystyle{IEEEtran}
\bibliography{biblio} 
\end{document}